\documentclass[journal]{IEEEtran}

\usepackage[T1]{fontenc}
\usepackage{amsmath,amssymb,amsfonts}
\usepackage{graphicx}
\usepackage{booktabs}
\usepackage{cite}
\usepackage{url}
\usepackage{bm}
\usepackage[hidelinks]{hyperref}

\usepackage{orcidlink}
\usepackage{fancyhdr}
\usepackage{multirow}
\usepackage{subcaption}
\usepackage{balance}
\newcommand{\Rhat}{\hat{R}}
\newcommand{\Rraw}{\hat{R}_0}
\newcommand{\Lhat}{\hat{L}}
\newcommand{\Leff}{\tilde{L}}

\begin{document}

\title{
Bright-Channel Retinex Enhancement with a Conditional Overdispersed-Noise Analysis
}

\author{
		Jongpil Jeong\orcidlink{0009-0003-5434-1601}~\IEEEmembership{Member,~IEEE}
		\thanks{
				J. Jeong is with the 3D Optical Imaging Systems Lab,
				Department of Eletronics and Information Communication Engineering,
				Faculty of Computer Science and Systems Engineering,
				Kyushu Institute of Technology, Iizuka, Fukuoka, Japan
				% and also with
				% Plenoptix, Busan, Republic of Korea
				(e-mail: \href{mailto:jongpil.jeong@ieee.org}{jongpil.jeong@ieee.org}). 
				% Code will be made publicly available at \href{https://github.com/Jongpil0911/BRINEX}{github.com/Jongpil0911/BRINEX} upon publication.
				}
}

\maketitle

% ─────────────────────────────────────────────────────────────────────────────
\begin{abstract}
I present a training-free low-light enhancement method that combines local bright-channel illumination estimation, Retinex division, and edge-preserving denoising.
For a fixed illumination estimate, a conditional Negative-Binomial pseudo-count model characterises the heteroscedastic noise amplified by division.
The unconstrained reflectance ratio is the pixelwise maximum-likelihood estimate, with a boundary solution for zero-valued observations; the implemented estimate additionally applies illumination flooring and range clipping.
The NB model is a diagnostic noise analysis rather than a calibrated sensor model, and the final fixed-bandwidth bilateral filter is an empirical approximation rather than the exact Bayesian solution.
On the LOL-v1 dataset, the method obtains mean PSNR/SSIM of 17.74\,dB/0.739, the highest values among the evaluated with conventional methods.
A $400\times600$ image is processed at approximately 43\,FPS on an Apple M2 Pro CPU.
\end{abstract}


% ─────────────────────────────────────────────────────────────────────────────
\section{Introduction}
\label{sec:intro}

Capturing images in low-light conditions is challenging because photon starvation amplifies signal-dependent shot noise and sensor read noise.
Retinex theory \cite{land1977retinex} models an observed image $I$ as:
\begin{equation}
	I(x) = R(x) \cdot L(x), \quad \forall x,
	\label{eq:retinex}
\end{equation}
where $R$ and $L$ represent the reflectance component (object albedo) and the illumination component, respectively.
Additionally, $x$ represents the spatial coordinates of the pixel.
Enhancing a low-light image then reduces to recovering $R$, which captures scene content independently of illumination conditions.

The primary problem is estimating $L$ without supervision.
Conventional image processing methods such as gamma correction (power law) and Contrast Limited Adaptive Histogram Equalization (CLAHE)~\cite{gonzalez2009digital}, Naturalness Preserved Enhancement (NPE)~\cite{npe2013}, and Low-light Image Enhancement (LIME)~\cite{lime2016} use statistical approaches but remain sensitive to illumination-estimation errors and post-division noise.
While many recent methods have turned to deep learning models \cite{retinexnet2018,zeroDCE2020,snraware2022,retinexformer2023} to circumvent these limitations, such data-driven approaches require extensive paired or unpaired datasets and often lack physical interpretability.

In this letter, I present BRINEX, a conventional bright-channel Retinex method accompanied by a conditional overdispersed-noise analysis.
Bright-channel features have also appeared in deep-learning models~\cite{lightennet2018,mbllen2018}; here, the local maximum is used directly to construct a standalone illumination estimate.
This work concentrates on how a conditional overdispersed-count model propagates through the subsequent Retinex illumination-division step.
Three directly testable insights follow:

\begin{enumerate}
	\item	\textbf{Conditional reflectance-domain noise model.}
			For a fixed illumination estimate, I derive the mean and variance of the scaled Negative Binomial (NB) count variable.
			The photon shot-noise term grows as illumination decreases, explaining the strong heteroscedasticity after division.

	\item	\textbf{Conditional MLE and regularised refinement.}
			I show that the Retinex division ($\hat{R}=I/\hat{L}$) is the pixelwise Maximum Likelihood Estimator (MLE) under the conditional NB likelihood.
			This does not preclude useful priors; empirically, however, the tested TV-regularised objectives reduce PSNR.

	\item 	\textbf{Empirical edge-preserving denoising.}
			The spatially varying variance motivates testing an edge-preserving denoiser rather than uniform smoothing.
			On Eval15, a fixed-bandwidth bilateral filter obtains the highest mean PSNR and SSIM among the tested Gaussian, Non-Local Means (NLM), and Guided Filter alternatives.
\end{enumerate}

% The complete BRINEX runs at ${\approx}43$\,FPS for $400\times600$ images on CPU, achieving PSNR = 17.74 dB / SSIM = 0.739 on LOL-v1, and serves as a compact, reproducible conventional image processing method.

The remainder of this letter is organised as follows.
Section~\ref{sec:background} describes the bright-channel Illumination Estimation Map (IEM) and the Negative Binomial statistics.
Section~\ref{sec:noise} derives the conditional noise statistics of the reflectance map and establishes its MLE interpretation.
Section~\ref{sec:method} describes the complete BRINEX.
Section~\ref{sec:experiments} presents quantitative and qualitative comparisons on the LOw-Light dataset (LOL)~\cite{retinexnet2018}.
Finally, I conclude this letter with Section~\ref{sec:conclusion}.

% ─────────────────────────────────────────────────────────────────────────────
\section{Background}
\label{sec:background}

\subsection{Illumination Estimation Map}
\label{ssec:bc_illum}

The implementation is motivated by the DCP transmission operator associated with the atmospheric scattering model~\cite{israel1959koschmieders,he2010single},
\begin{equation}
 I(x)=J(x)t(x)+A\bigl(1-t(x)\bigr),
 \label{eq:asm}
\end{equation}
where $A$ and $t$ denote atmospheric light and transmission, respectively. BRINEX uses this formulation only as an operator-level construction for detecting insufficient illumination; it does not assume that a low-light image is physically generated by atmospheric scattering and does not recover scene depth.

For a low-light RGB image $I\in[0,1]^{H\times W\times3}$, I compute the patch-wise \emph{Bright Channel} over a $15\times15$ window $\Omega(x)$:
\begin{equation}
 J^{\mathrm{bright}}(x)=\max_{p\in\Omega(x)}\left(\max_{c\in\{r,g,b\}} I^c(p)\right).
 \label{eq:jbright}
\end{equation}
If this local maximum is small, no channel in the neighbourhood contains a sufficiently bright pixel; $J^{\mathrm{bright}}(x)\ll1$ therefore identifies an under-exposed region and serves as a local illumination proxy~\cite{dong2011bcp}.
The coarse \emph{illumination estimation map} (IEM) is expressed as:
\begin{equation}
 \hat{L}(x)=\omega\cdot J^{\mathrm{bright}}(x),\qquad\omega=0.95.
 \label{eq:coarse_illum}
\end{equation}
The implementation evaluates the algebraically related complement-domain dark-channel form and includes per-channel normalisation to reduce colour imbalance~\cite{he2010single}; this is a computational construction of the IEM, not an atmospheric-scattering or depth-recovery model. I use a guided filter~\cite{he2010guided} with grayscale guidance (radius 60, regularisation $10^{-4}$) to refine $\hat{L}$. I define the effective illumination $\Leff(x)=\max\{\Lhat(x),\varepsilon\}$ with $\varepsilon=0.05$, and project the divided RGB values onto the valid range:
\begin{equation}
		\hat{R}^{c}(x) =
		\Pi_{[0,1]}\!\left(\frac{I^{c}(x)}{\Leff(x)}\right),
		\qquad c\in\{r,g,b\}.
		\label{eq:Rhat}
\end{equation}

\subsection{Negative Binomial Photon Statistics}
\label{ssec:nb}

Photon counts from thermal or partially coherent light can exhibit bunching and may be described by a Negative Binomial (NB) distribution \cite{goodman2015statistical}, defined as:
\begin{equation}
		P(n;\,\mu,r) = \binom{n+r-1}{n}
									 \!\left(\frac{r}{\mu+r}\right)^{\!r}
									 \!\left(\frac{\mu}{\mu+r}\right)^{\!n},
		\label{eq:nb}
\end{equation}
where $n$ is the photon count, $\mu$ is the mean photon count, and
$r > 0$ is the overdispersion parameter. The variance is explicitly given by:
\begin{equation}
		\sigma^2 = \mu + \frac{\mu^2}{r}.
		\label{eq:nb_var}
\end{equation}
The moment-generating function of NB($\mu, r$) satisfies:
\begin{equation}
	\lim_{r\rightarrow \infty}\underbrace{\left(1 + {\frac{\mu(1-e^t)}{r}}\right)^{-r}}_{\text{NB MGF}} = e^{\mu(e^t - 1)} = \underbrace{M_{\mathrm{Pois}(\mu)}(t)}_{\text{Poisson MGF}},
\end{equation}
which, by the L\'{e}vy continuity theorem, implies distributional convergence to Poisson($\mu$), with $\sigma^2 \rightarrow \mu$ as a consequence.
Finite $r$ represents super-Poissonian dispersion ($\sigma^2>\mu$).

Poisson noise assumes independent photon arrivals and predicts $\sigma^2=\mu$. 
Processed camera images, however, undergo demosaicing, gain amplification, nonlinear colour processing, and quantisation. 
I therefore use the NB law only as a phenomenological overdispersion model. 
The LOL images are not RAW measurements, $g=255$ is a digital scaling factor rather than a calibrated sensor gain, and $r$ is neither calibrated nor used by the fixed-bandwidth enhancement. 
Consequently, the following analysis is a conditional diagnostic model, not evidence that the stored RGB values are physical photon counts.

% ─────────────────────────────────────────────────────────────────────────────
\section{Noise Analysis in Reflectance Domain}
\label{sec:noise}

\subsection{Conditional Scaled-NB Count Model}
\label{ssec:scaled_nb}

The effective illumination $\Leff(x)$ from Section~\ref{ssec:bc_illum} defines the unclipped ratio $\Rraw^c=I^c/\Leff$ and the implemented estimate $\Rhat^c=\Pi_{[0,1]}(\Rraw^c)$.
For a conditional diagnostic analysis, an 8-bit channel value is represented as the pseudo-count $K^c(x)=gI^c(x)$, where $g=255$, $K^c(x)\in\{0,\ldots,255\}$, and $I^c(x)=K^c(x)/g$.

Holding $\Leff$ fixed, the Retinex factorisation is encoded channel-wise in $\mu^c(x)=gR^c(x)\Leff(x)$, giving:
\begin{equation}
\begin{aligned}
 K^c(x)\mid R^c,\Leff &\sim
 \mathrm{NB}\!\left(\mu^c(x)=gR^c(x)\Leff(x),r\right),\\
 I^c(x)&=K^c(x)/g.
\end{aligned}
 \label{eq:count_model}
\end{equation}
Assuming $\Leff$ equals the true illumination, the conditional moments of the unclipped ratio are
\begin{align}
 \mathbb{E}[\Rraw^c(x)\mid R^c,\Leff] &= R^c(x), \label{eq:mean_rhat}\\
 \mathrm{Var}[\Rraw^c(x)\mid R^c,\Leff]
 &=\underbrace{\frac{R^c(x)}{g\Leff(x)}}_{\text{shot noise}}+\underbrace{\frac{R^c(x)^2}{r}}_{\text{excess dispersion}}.
 \label{eq:var_rhat}
\end{align}
The first term grows as $\Leff(x)$ decreases, illustrating heteroscedasticity after Retinex division. Illumination-estimation errors, clipping, spatial correlations, read noise, and camera processing introduce bias and variance outside this idealised model.

\begin{figure}[!h]
	\centering
	\includegraphics[width=0.78\columnwidth]{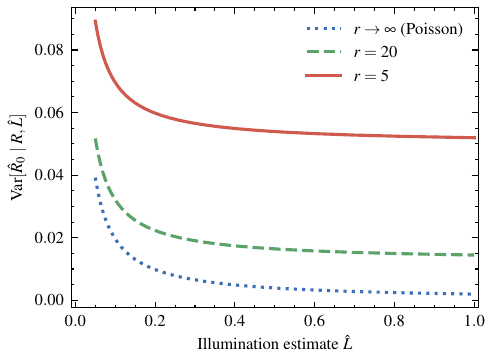}
	\caption{Conditional variance from Eq.~\eqref{eq:var_rhat} versus
	illumination for illustrative $R=0.5$ and $g=255$. Lower illumination
	amplifies the shot-noise term; finite $r$ adds excess dispersion.}
	\label{fig:variance}
\end{figure}

\subsection{Conditional MLE and Regularised Refinement}
\label{ssec:map_opt}

Consider the NB-MAP objective in the reflectance domain:
\begin{equation}
	\hat{R}_\mathrm{MAP} = \arg\min_{R}
	\underbrace{\sum_x \mathcal{L}_\mathrm{NB}
	(gR(x)\Leff(x),\,K(x))}_{\text{NB-NLL}}
	+ \alpha\,\mathrm{TV}(R),
	\label{eq:map}
\end{equation}
where $\mathcal{L}_\mathrm{NB}$ is the NB negative log-likelihood and $\mathrm{TV}$ denotes isotropic total variation. The gradient of the data term with respect to $R(x)$ is
\begin{equation}
	\frac{\partial \mathcal{L}_\mathrm{NB}}{\partial R(x)} =
	\frac{rg\Leff(x)\bigl(R(x)-\Rraw(x)\bigr)}
	{R(x)\bigl(gR(x)\Leff(x)+r\bigr)},
	\label{eq:grad_nb}
\end{equation}
For $K(x)>0$, this derivative vanishes at the unique interior optimum $\Rraw(x)=I(x)/\Leff(x)$. For $K(x)=0$, the likelihood is minimised at the boundary $R(x)=0$. Thus the constrained conditional pixelwise MLE is the projection in Eq.~\eqref{eq:Rhat}; it is biased relative to Eq.~\eqref{eq:mean_rhat} wherever upper clipping is active. The ratio MLE follows from the mean parameterisation $\mu=gR\Leff$, rather than specifically from the NB distribution. The NB-specific content is its conditional variance in Eq.~\eqref{eq:var_rhat}. For $\alpha>0$, Eq.~\eqref{eq:map} defines a different MAP estimator that trades likelihood against spatial regularity; the tested TV settings reduce mean PSNR relative to $\Rhat$ (Section~\ref{sec:experiments}).

% ─────────────────────────────────────────────────────────────────────────────
\section{Proposed Method}
\label{sec:method}

\begin{figure*}[t]
\centering
\newcommand{\qw}{0.119\textwidth}
% ── Row 1: Image 111 ──────────────────────────────────────────────────────────
\begin{subfigure}{\qw}\centering
    \includegraphics[width=\textwidth]{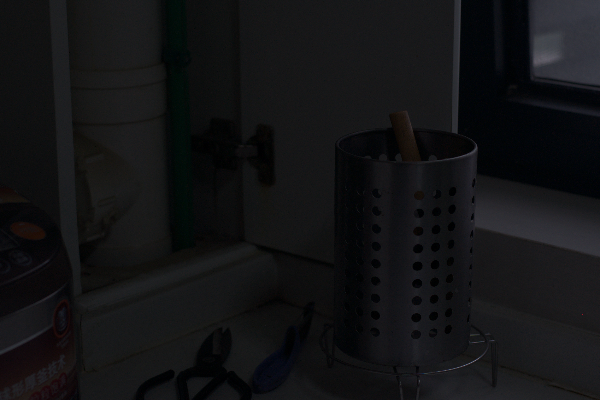}
    \caption*{{\scriptsize 5.61\,dB\,/\,0.187}}\end{subfigure}\hfill
\begin{subfigure}{\qw}\centering
    \includegraphics[width=\textwidth]{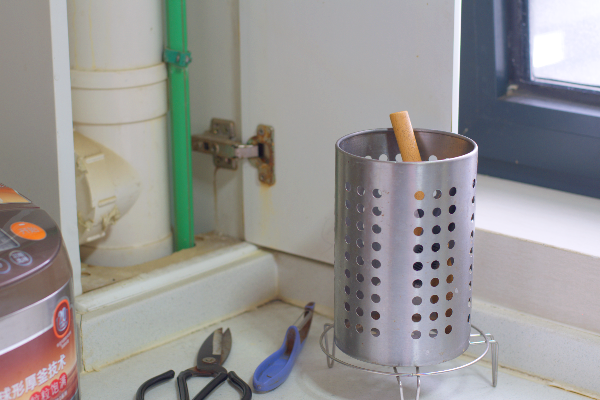}
    \caption*{}\end{subfigure}\hfill
\begin{subfigure}{\qw}\centering
    \includegraphics[width=\textwidth]{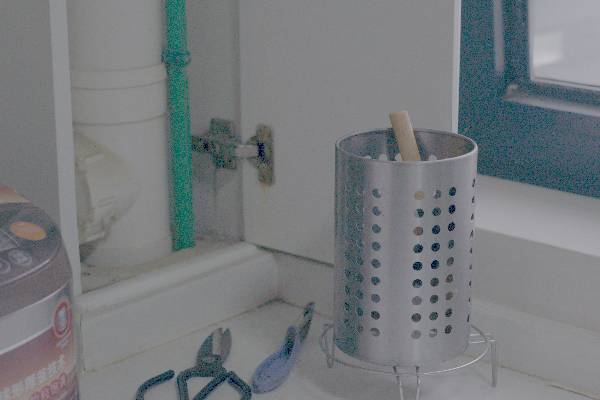}
    \caption*{{\scriptsize 17.00\,dB\,/\,0.753}}\end{subfigure}\hfill
\begin{subfigure}{\qw}\centering
    \includegraphics[width=\textwidth]{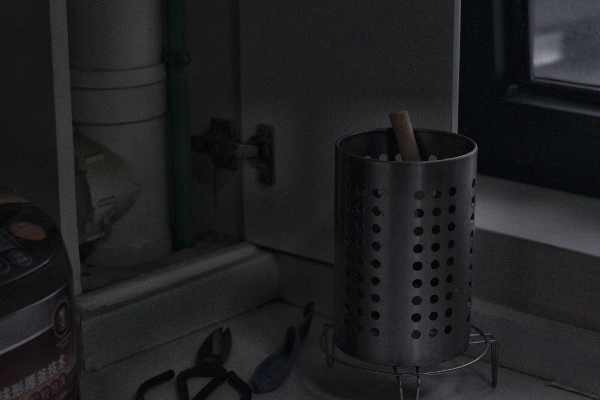}
    \caption*{{\scriptsize 6.48\,dB\,/\,0.308}}\end{subfigure}\hfill
\begin{subfigure}{\qw}\centering
    \includegraphics[width=\textwidth]{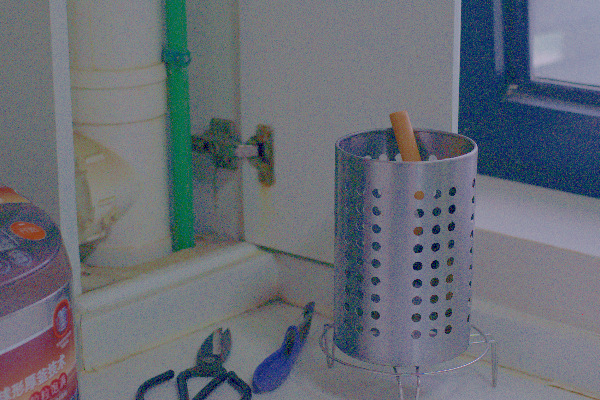}
    \caption*{{\scriptsize 14.98\,dB\,/\,0.537}}\end{subfigure}\hfill
\begin{subfigure}{\qw}\centering
    \includegraphics[width=\textwidth]{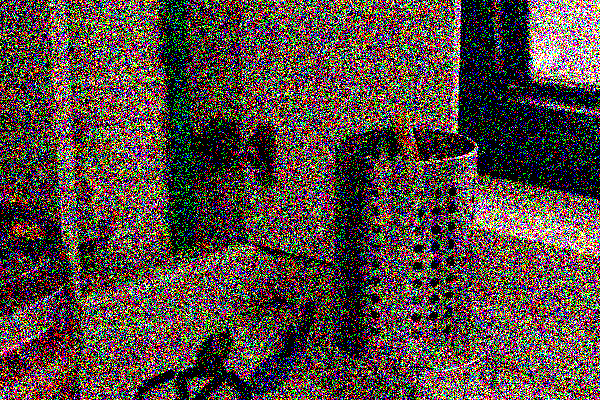}
    \caption*{{\scriptsize 5.60\,dB\,/\,0.019}}\end{subfigure}\hfill
\begin{subfigure}{\qw}\centering
    \includegraphics[width=\textwidth]{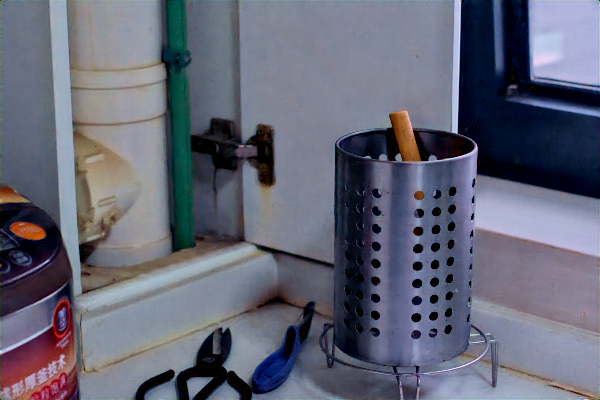}
    \caption*{{\scriptsize 14.68\,dB\,/\,0.760}}\end{subfigure}\hfill
\begin{subfigure}{\qw}\centering
    \includegraphics[width=\textwidth]{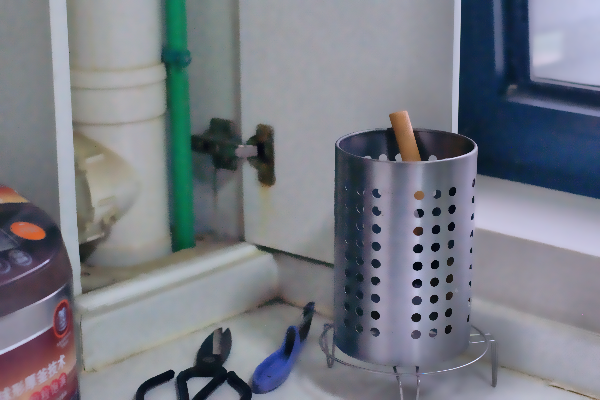}
    \caption*{{\scriptsize 18.53\,dB\,/\,0.861}}\end{subfigure}
\vspace{0.5mm}
% ── Row 2: Image 146 ──────────────────────────────────────────────────────────
\begin{subfigure}{\qw}\centering
    \includegraphics[width=\textwidth]{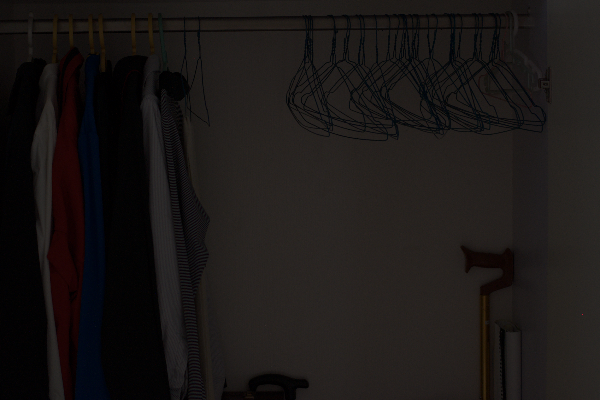}
    \caption*{{\scriptsize 6.63\,dB\,/\,0.260}}\end{subfigure}\hfill
\begin{subfigure}{\qw}\centering
    \includegraphics[width=\textwidth]{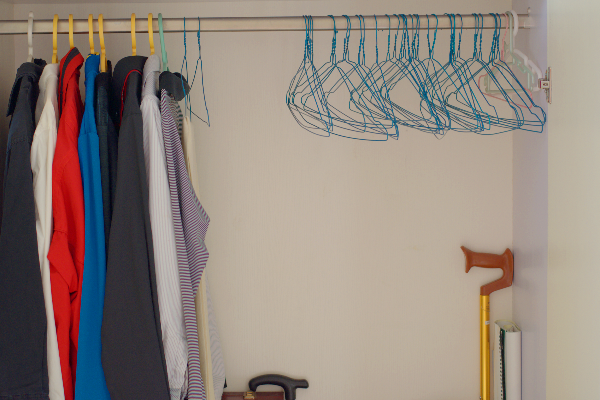}
    \caption*{}\end{subfigure}\hfill
\begin{subfigure}{\qw}\centering
    \includegraphics[width=\textwidth]{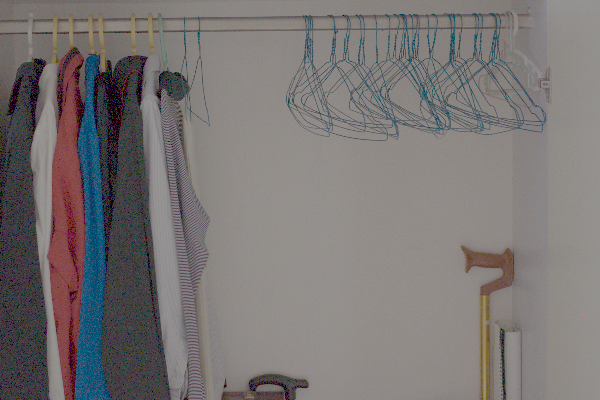}
    \caption*{{\scriptsize 18.74\,dB\,/\,0.776}}\end{subfigure}\hfill
\begin{subfigure}{\qw}\centering
    \includegraphics[width=\textwidth]{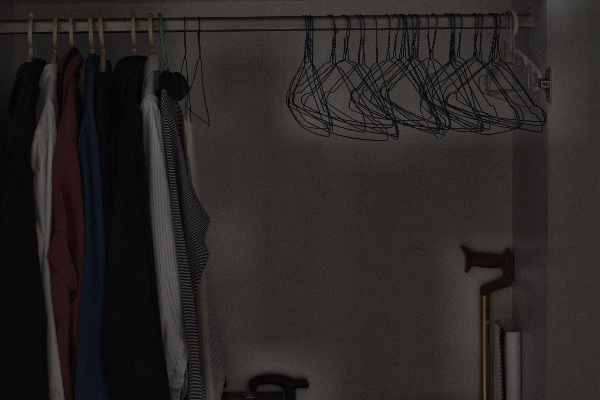}
    \caption*{{\scriptsize 7.73\,dB\,/\,0.409}}\end{subfigure}\hfill
\begin{subfigure}{\qw}\centering
    \includegraphics[width=\textwidth]{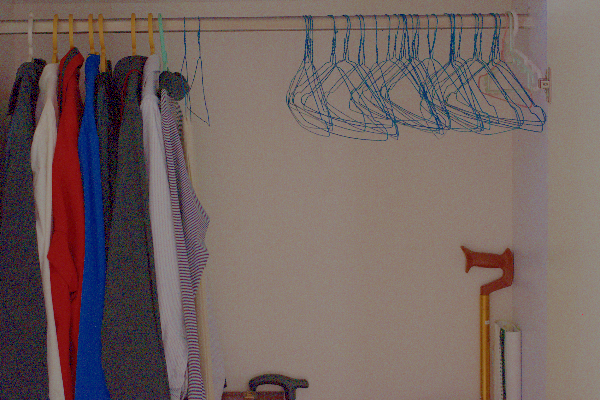}
    \caption*{{\scriptsize 16.36\,dB\,/\,0.662}}\end{subfigure}\hfill
\begin{subfigure}{\qw}\centering
    \includegraphics[width=\textwidth]{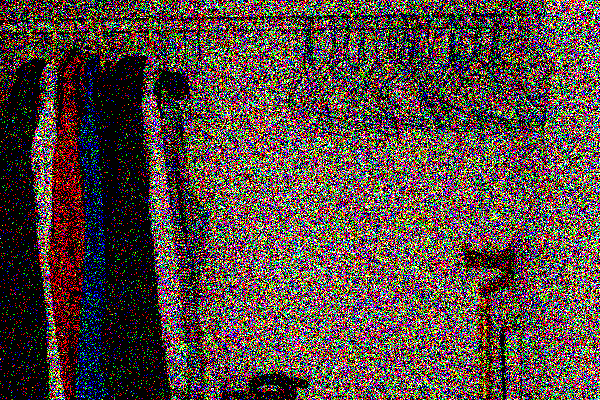}
    \caption*{{\scriptsize 6.07\,dB\,/\,0.033}}\end{subfigure}\hfill
\begin{subfigure}{\qw}\centering
    \includegraphics[width=\textwidth]{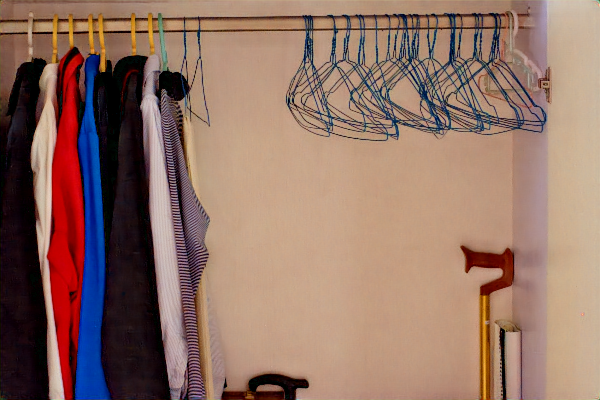}
    \caption*{{\scriptsize 20.50\,dB\,/\,0.794}}\end{subfigure}\hfill
\begin{subfigure}{\qw}\centering
    \includegraphics[width=\textwidth]{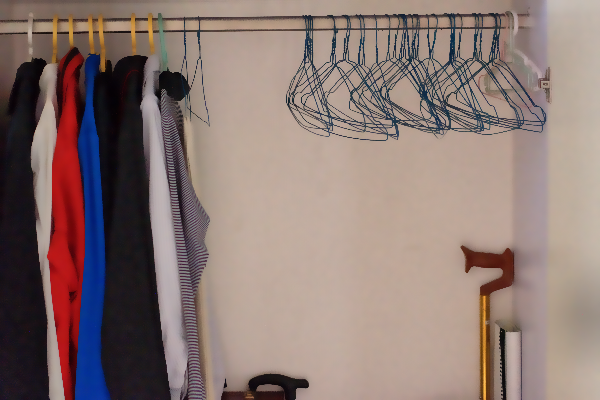}
    \caption*{{\scriptsize 23.66\,dB\,/\,0.863}}\end{subfigure}
\vspace{0.5mm}
% ── Row 3: Image 669 ──────────────────────────────────────────────────────────
\begin{subfigure}{\qw}\centering
    \includegraphics[width=\textwidth]{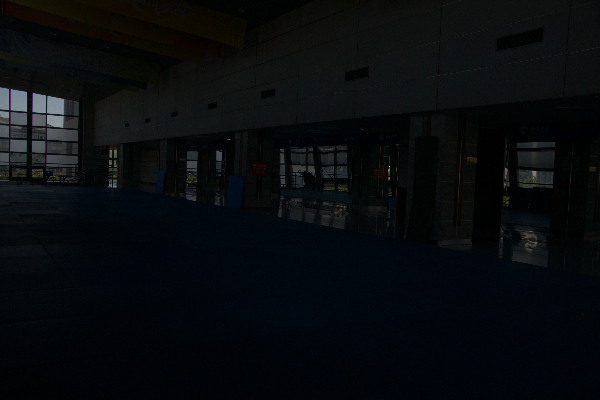}
    \caption*{{\scriptsize 12.12\,dB\,/\,0.295} \centering Input}\end{subfigure}\hfill
\begin{subfigure}{\qw}\centering
    \includegraphics[width=\textwidth]{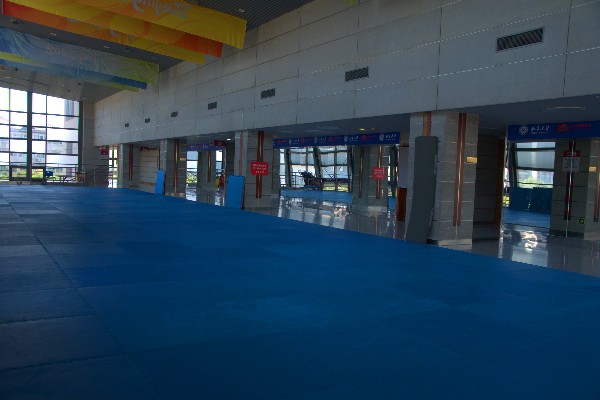}
    \caption*{\\ \centering GT}\end{subfigure}\hfill
\begin{subfigure}{\qw}\centering
    \includegraphics[width=\textwidth]{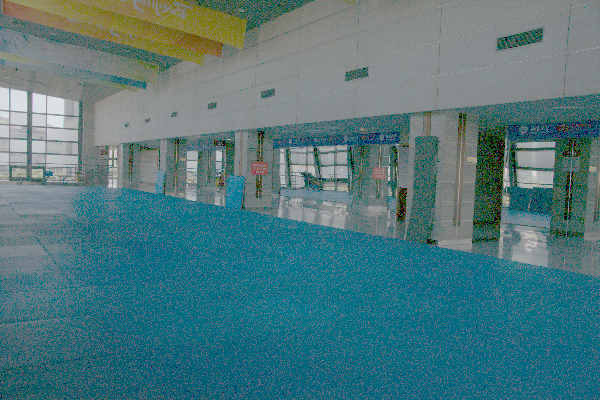}
    \caption*{{\scriptsize 14.09\,dB\,/\,0.520} \centering Gamma}\end{subfigure}\hfill
\begin{subfigure}{\qw}\centering
    \includegraphics[width=\textwidth]{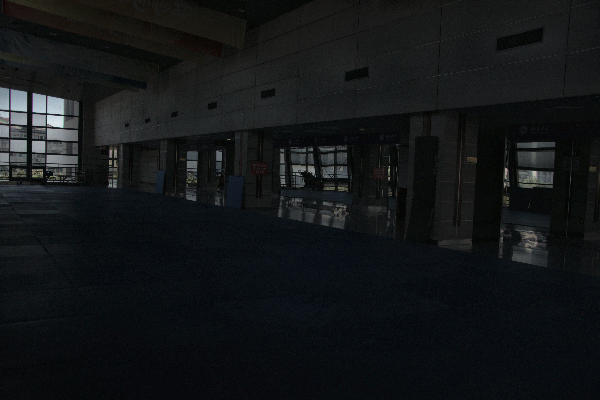}
    \caption*{{\scriptsize 13.77\,dB\,/\,0.409} \centering CLAHE}\end{subfigure}\hfill
\begin{subfigure}{\qw}\centering
    \includegraphics[width=\textwidth]{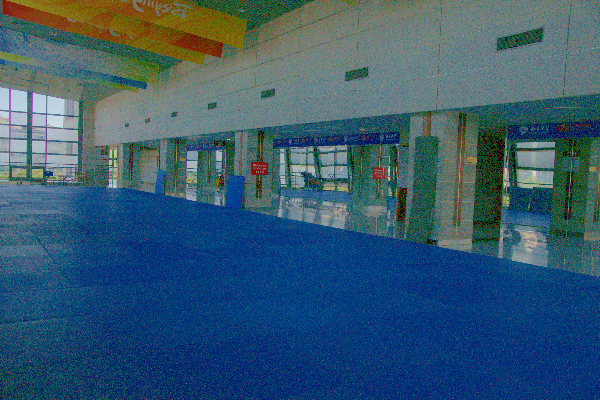}
    \caption*{{\scriptsize 18.57\,dB\,/\,0.426} \centering NPE}\end{subfigure}\hfill
\begin{subfigure}{\qw}\centering
    \includegraphics[width=\textwidth]{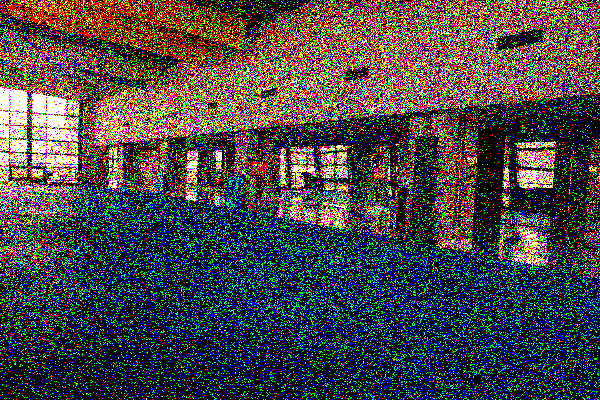}
    \caption*{{\scriptsize 7.97\,dB\,/\,0.067} \centering PCD}\end{subfigure}\hfill
\begin{subfigure}{\qw}\centering
    \includegraphics[width=\textwidth]{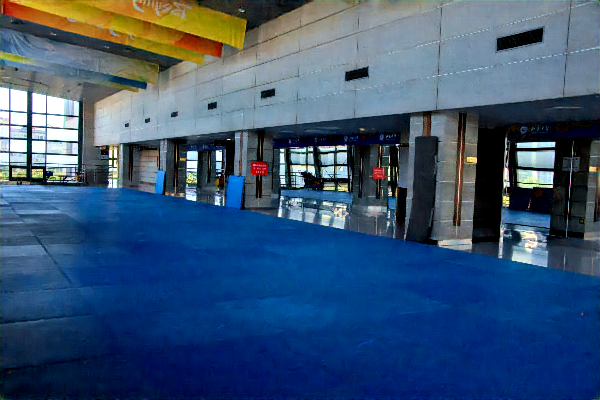}
    \caption*{{\scriptsize 17.06\,dB\,/\,0.734} \centering MBLLEN}\end{subfigure}\hfill
\begin{subfigure}{\qw}\centering
    \includegraphics[width=\textwidth]{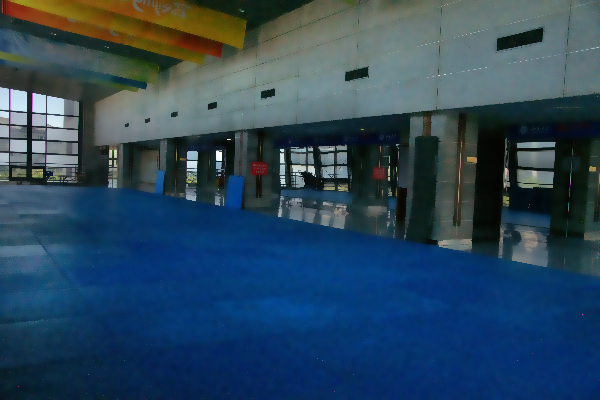}
    \caption*{{\scriptsize 21.57\,dB\,/\,0.757} \centering Ours-B}\end{subfigure}
	\vspace{1mm}
	\caption{Qualitative comparison on three eval15 images.
	Columns: Input, GT, Gamma ($\gamma{=}0.23$), CLAHE, NPE, PCD, MBLLEN (trained), and Ours-B.
	PSNR\,(dB)\,/\,SSIM shown below each result.}
	\label{fig:qualitative}
\end{figure*}

\subsection{Bilateral Post-Processing}
\label{ssec:bilateral}

Equation~\eqref{eq:var_rhat} motivates noise suppression in locally smooth regions while avoiding uniform smoothing across reflectance boundaries. I therefore evaluate the bilateral filter~\cite{tomasi1998bilateral},
\begin{equation}
	\tilde{R}(x) = \frac{1}{Z(x)}\sum_{q\in\Omega(x)}
		k_s(\|x-q\|)\,k_r(\Rhat(x)-\Rhat(q))\,\Rhat(q),
	\label{eq:bilateral}
\end{equation}
where $k_s(d)=e^{-d^2/(2\sigma_s^2)}$, $k_r(\delta)=e^{-\delta^2/(2\sigma_c^2)}$, and $Z(x)$ is the normalising constant. In flat regions the filter smooths neighbouring estimates, whereas its range kernel limits averaging across strong reflectance differences.
The standard bilateral filter is not the exact MAP estimator of Eq.~\eqref{eq:map}. To bridge this gap, I take a second-order Taylor expansion of the NB log-likelihood around the MLE $\Rhat$. Differentiating Eq.~\eqref{eq:grad_nb} with respect to $R$ at $R=\Rraw$ gives the observed Fisher information:
\begin{equation}
	\left.\frac{\partial^2 \mathcal{L}_\mathrm{NB}}{\partial R(x)^2}\right|_{R=\Rraw} = \frac{rg\Leff(x)}{\Rraw(x)\bigl(g\Rraw(x)\Leff(x)+r\bigr)}.
	\label{eq:fisher_info}
\end{equation}
Algebraic rearrangement of Eq.~\eqref{eq:var_rhat} shows that this second derivative equals the reciprocal of the conditional variance, i.e., $1/\sigma^2(\Rraw)$. Since the first derivative is zero at the MLE, the Taylor expansion locally approximates the MAP objective as a heteroscedastic weighted least-squares problem:
\begin{equation}
	\hat{R}_\mathrm{MAP} \approx \arg\min_{R} \sum_x \frac{(R(x) - \Rhat(x))^2}{2\sigma^2(x)} + \alpha\,\mathrm{TV}(R).
	\label{eq:taylor_map}
\end{equation}
Replacing TV locally by a robust pairwise surrogate with bilateral weights $w_{xq}=k_s(\|x-q\|)k_r(\Rhat(x)-\Rhat(q))$~\cite{elad2002origin,milanfar2013tour}, a single Jacobi update initialised at $R^{(0)}=\Rhat$ has the form
\begin{equation}
	R^{(1)}(x) = \frac{\dfrac{\hat{R}(x)}{\sigma^2(x)} + \alpha \sum_q w_{xq} \hat{R}(q)}{\dfrac{1}{\sigma^2(x)} + \alpha \sum_q w_{xq}}
\end{equation}
This update differs from the standard bilateral filter through the data-fidelity term $1/\sigma^2(x)$ and coincides with it only in the limiting case where that term vanishes. I therefore do not identify the implemented bilateral filter with the conditional NB-MAP solution. Instead, I use it as a fast empirical edge-preserving post-processor and report in Section~\ref{ssec:ablation} that the tested variance-adaptive variants did not improve Eval15 accuracy.
I process 8-bit RGB reflectance values with neighbourhood diameter $d=9$, $\sigma_s=75$ pixels, and $\sigma_c=75$ intensity levels. Because $\sigma_s$ is much larger than the $9\times9$ support, the spatial weights are nearly uniform within that support; the practical selectivity comes primarily from the range kernel.
The complete pipeline has no learned parameters: given $I$, compute $\hat{R}$ via~\eqref{eq:Rhat}, and apply bilateral filtering with $d{=}9$, $\sigma_s{=}75$, $\sigma_c{=}75$. The numerical hyperparameters are fixed across all reported images.

It is important to distinguish the diagnostic model from the implemented enhancement algorithm. Neither $r$ nor the conditional variance in Eq.~\eqref{eq:var_rhat} is used to determine the output or the fixed bilateral-filter parameters. BRINEX is therefore a bright-channel Retinex pipeline accompanied by an NB noise analysis, rather than an NB-driven estimator.

% ─────────────────────────────────────────────────────────────────────────────
\section{Experimental Setup and Results}
\label{sec:experiments}

\subsection{Dataset and Specification}
\label{ssec:dataset}

I evaluate on the \textbf{LOL-v1} dataset~\cite{retinexnet2018}, which provides paired low/normal-light images; \emph{Eval15} contains 15 test pairs. Performance is measured by PSNR and SSIM (scikit-image), FSIMc~\cite{zhang2011fsim} (\texttt{piq}), and a custom implementation of TMQI~\cite{yeganeh2013tmqi}, using 8-bit RGB outputs and paired references. TMQI is secondary because it was designed for tone-mapped HDR imagery. Quality metrics were recomputed from saved outputs with the same evaluation script. BRINEX runtime was measured with Python 3.11, OpenCV, and NumPy on an Apple M2 Pro; reported means over 15 images should not be interpreted as confidence bounds or broad cross-dataset generalisation.

% \begin{figure}[!h]
% \centering
% \begin{subfigure}{0.22\textwidth}
%     \centering
%     \includegraphics[width=\textwidth]{images/06_low_111.png}
%     \caption*{{\scriptsize Input\\5.61\,dB\,/\,0.187}}
% \end{subfigure}
% \begin{subfigure}{0.22\textwidth}
%     \centering
%     \includegraphics[width=\textwidth]{images/03_high_111.png}
%     \caption*{{\scriptsize GT}}
% \end{subfigure}
% \begin{subfigure}{0.22\textwidth}
%     \centering
%     \includegraphics[width=\textwidth]{images/02_gamma_tf_03_111.png}
%     \caption*{{\scriptsize Gamma\\17.00\,dB\,/\,0.753}}
% \end{subfigure}
% \begin{subfigure}{0.22\textwidth}
%     \centering
%     \includegraphics[width=\textwidth]{images/10_ours_b_03_111.png}
%     \caption*{{\scriptsize Ours-B\\18.53\,dB\,/\,0.861}}
% \end{subfigure}
% 	\caption{Qualitative comparison (an eval15 image). Gamma ($\gamma{=}0.23$) represents the classical baselines of Table~\ref{tab:sota}; PSNR/SSIM below each result.}
% 	\label{fig:qualitative}
% \end{figure}

\subsection{Qualitative Evaluations}
\label{ssec:qualitative}

Figure~\ref{fig:qualitative} compares an eval15 case against gamma correction ($\gamma=0.23$), representative of the classical controls in Table~\ref{tab:sota}: it brightens the input however retains poor visibility, whereas the proposed output suppresses flat-region noise and better recovers boundaries, consistent with its quantitative advantage.

\subsection{Quantitative Evaluations}
\label{ssec:quantitative}

Table~\ref{tab:sota} compares mean Eval15 scores against image processing methods and trained baselines.
\textbf{Ours-B} obtains the highest mean PSNR (17.74\,dB), SSIM (0.739), and FSIMc (0.918) among the evaluated non-trained methods; \textbf{Ours-A} attains the highest non-trained TMQI (0.8917, vs.\ 0.8909 for PhotonMLE). Bilateral post-processing raises the former fidelity metrics but lowers TMQI from 0.892 to 0.866, indicating a non-negligible naturalness/detail trade-off rather than an unqualified improvement. Among all methods, MBLLEN attains the highest TMQI overall (0.927), while Retinexformer remains substantially stronger in PSNR/SSIM/FSIMc.

\begin{table}[t]
\centering
\caption{Quantitative comparison on LOL-v1 with run-time.}
\label{tab:sota}
\setlength{\tabcolsep}{2pt}
\renewcommand{\arraystretch}{0.92}
\resizebox{\columnwidth}{!}{%
\scriptsize
\begin{tabular}{lcccc|cc}
\toprule
\multirow[c]{2}{*}{Method}
& PSNR [$\uparrow$]
& \multirow[c]{2}{*}{SSIM [$\uparrow$]}
& \multirow[c]{2}{*}{FSIMc [$\uparrow$]}
& \multirow[c]{2}{*}{TMQI [$\uparrow$]}
& \multicolumn{2}{c}{FPS} \\
% \cmidrule(lr){6-7}
& [dB] & & & &{\tiny CPU} & {\tiny GPU}\\
\midrule
\midrule
\multicolumn{7}{l}{\textit{Image processing method}} \\
Gamma TF ($\gamma{=}0.23$)                      & 16.73 & 0.630 & 0.871 & 0.867 & 107 & 2462 \\
CLAHE (clip\,2.0, $15{\times}15$)               &  8.90 & 0.337 & 0.799 & 0.734 & 112 & ---- \\
NPE \cite{npe2013}\textsuperscript{\S}          & 16.59 & 0.487 & 0.872 & 0.864 & 127 & 594 \\
LIME \cite{lime2016}\textsuperscript{\S}        & 10.58 & 0.386 & 0.779 & 0.771 & 0.02 & ---- \\
PhotonMLE$^\ddagger$ \cite{cho2016peplography}  & 17.11 & 0.493 & 0.907 & 0.891 & 0.3 & ---- \\
Kalman$^\ddagger$ \cite{kim2023kalman}          & 16.18 & 0.634 & 0.867 & 0.829 & 17 & ---- \\
Ours-A (BC$\to$IEM, MLE div.)                   & 17.17 & 0.496 & 0.908 & \textbf{0.892} & 42 & 319 \\
\textbf{Ours-B} (BC$\to$IEM\,+\,Bil.)           & \textbf{17.74} & \textbf{0.739} & \textbf{0.918} & 0.866 & 43 & 271 \\
\midrule
\multicolumn{7}{l}{\textit{Learning-based method}} \\
Zero-DCE \cite{zeroDCE2020}$^\ddagger$                        & 14.80 & 0.561 & 0.918 & 0.830 & 3.0 & 68 \\
Retinex-Net \cite{retinexnet2018}$^\ddagger$                  & 16.70 & 0.424 & 0.846 & 0.893 & 1.7 & 22.9 \\
MBLLEN \cite{mbllen2018}$^\ddagger$                           & 17.47 & 0.726 & 0.882 & \underline{\textbf{0.927}} & 0.49 & 9.4 \\
\underline{Retinexformer} \cite{retinexformer2023}$^\dagger$  & \underline{\textbf{25.15}} & \underline{\textbf{0.843}} & \underline{\textbf{0.961}} & 0.881 & ---- & 53 \\
\bottomrule
\end{tabular}%
}% end resizebox
\vspace{2pt}
\parbox{\columnwidth}{\scriptsize
$^\S$Reimplementation of the cited enhancement method.
$^\ddagger$Official weights where available or an adaptation/reimplementation evaluated from saved outputs.
$^\dagger$Official pretrained model.
FPS values are implementation- and device-dependent and are not a hardware-normalised ranking.}
\end{table}
% \subsection{Generalisation on LOL-v1 Full}
% \label{ssec:generalization}

% I report results on the 485-image subset, which is typically used as the \emph{training} set for supervised methods. For this training-free method it is a larger within-dataset stress test. The bilateral filter consistently improves over the MLE baseline (PSNR from 15.38\,dB to 15.81\,dB, $\Delta$+0.43\,dB; SSIM from 0.639 to 0.740, $\Delta$+0.101) across 485 diverse scene conditions, demonstrating generalisation that does not depend on any scene-specific tuning.

\subsection{Ablation Study}
\label{ssec:ablation}

\paragraph{Effect of TV-regularised refinement.}
I evaluate TV-regularised NB-MAP (ISTA, 100 iterations) for $\alpha\in\{0.005, 0.01, 0.02, 0.05, 0.1\}$. All settings produce PSNR $\leq 16.61$\,dB, below the conditional MLE baseline (17.17\,dB) -- plausibly because isotropic TV penalises reflectance gradients uniformly, including genuine edges the bilateral kernel instead preserves, and the ISTA step size was not jointly tuned with $\alpha$. This is specific to the tested objective, schedule, and $\alpha$ range, not a claim of general MLE--MAP equivalence.

\paragraph{Choice of spatial denoiser.}
Table~\ref{tab:ablation_denoiser} compares four edge-preserving post-processors applied to $\Rhat$. The Guided Filter \cite{he2010guided} uses the low-light image as a guidance signal, inadvertently propagating noise correlations from the input: it degrades PSNR by 0.40\,dB relative to the unfiltered MLE baseline and gives the smallest SSIM gain among the four post-processors. Non-Local Means (NLM) \cite{buades2005nlm} improves SSIM substantially however still falls 0.18\,dB short of the bilateral filter in PSNR. The bilateral filter achieves the best result on both metrics, consistent with the need to suppress noise without uniform cross-edge averaging.

\begin{table}[t]
\centering
\caption{Denoiser ablation on LOL-v1.}
\label{tab:ablation_denoiser}
\begin{tabular}{lcccc}
\toprule
Post-processor & PSNR$\uparrow$ & SSIM$\uparrow$ & $\Delta$PSNR & $\Delta$SSIM \\
\midrule
None (MLE baseline)  & 17.17 & 0.496 & -- & -- \\
Gaussian ($\sigma=3$)     & 17.30 & 0.652 & $+$0.12 & $+$0.156 \\
Guided Filter \cite{he2010guided} & 16.77 & 0.588 & $-$0.40 & $+$0.092 \\
NLM \cite{buades2005nlm}          & 17.55 & 0.727 & $+$0.38 & $+$0.231 \\
\textbf{Bilateral (proposed)} & \textbf{17.74} & \textbf{0.739} & $\bm{+}$\textbf{0.56} & $\bm{+}$\textbf{0.243} \\
\bottomrule
\end{tabular}
\end{table}

\paragraph{$\sigma_c$ sensitivity and spatial adaptivity.}
For 8-bit reflectance values, with $d=9$ and $\sigma_s=75$ fixed, PSNR varies by $<0.04$\,dB for $\sigma_c\in[50,125]$ (17.74\,dB/0.739 at $\sigma_c=75$). I additionally evaluated the update motivated by Eq.~\eqref{eq:taylor_map} and a variant with $\sigma_c(x)\propto\sigma(x)$, sweeping $r\in\{2,5,10,20,50\}$ against $\alpha\in\{0.01,\dots,50\}$ or $k\in\{0.5,\dots,2\}$ (60 settings). The best adaptive setting reached 17.62\,dB/0.729, below the fixed filter. This negative result shows that the tested NB-adaptive constructions do not explain the fixed filter's empirical advantage; it does not establish that spatial adaptivity is generally ineffective.

\paragraph{Runtime.}
The complete BC and bilateral pipeline runs at approximately 43\,FPS for $400\times600$ images on an Apple M2 Pro CPU (median over the timed eval15 subset; file I/O excluded), without model loading, GPU use, or test-time optimisation.

% ─────────────────────────────────────────────────────────────────────────────
\section{Conclusion}
\label{sec:conclusion}

I presented a training-free bright-channel Retinex pipeline with fixed bilateral post-processing. A conditional NB pseudo-count model shows how an overdispersed variance propagates through illumination division and yields the unclipped ratio as the pixelwise MLE, with the appropriate zero-count boundary case. On the 15-image LOL-v1 Eval15 set, the method obtains mean PSNR/SSIM of 17.74\,dB/0.739 at approximately 43\,FPS. The NB analysis is diagnostic and does not calibrate or determine the fixed bilateral parameters. Limitations include evaluation on a single small test set, illumination bias in bright or saturated regions, clipping, spatial dependence from camera processing, and the use of $g=255$ as a digital scale rather than a sensor gain.

% ─────────────────────────────────────────────────────────────────────────────
\balance
\bibliographystyle{IEEEtran}
\bibliography{288_refs}

@article{he2010guided,
  title     = {Guided image filtering},
  author    = {He, Kaiming and Sun, Jian and Tang, Xiaoou},
  journal   = {IEEE transactions on pattern analysis and machine intelligence},
  volume    = {35},
  number    = {6},
  pages     = {1397--1409},
  year      = {2012},
  publisher = {IEEE}
}

@inproceedings{buades2005nlm,
  title        = {A non-local algorithm for image denoising},
  author       = {Buades, Antoni and Coll, Bartomeu and Morel, J-M},
  booktitle    = {2005 IEEE computer society conference on computer vision and pattern recognition (CVPR'05)},
  volume       = {2},
  pages        = {60--65},
  year         = {2005},
  organization = {Ieee}
}

@article{land1977retinex,
  title     = {The retinex theory of color vision},
  author    = {Land, Edwin H},
  journal   = {Scientific american},
  volume    = {237},
  number    = {6},
  pages     = {108--129},
  year      = {1977},
  publisher = {JSTOR}
}

@inproceedings{tomasi1998bilateral,
  title        = {Bilateral filtering for gray and color images},
  author       = {Tomasi, Carlo and Manduchi, Roberto},
  booktitle    = {Sixth international conference on computer vision (IEEE Cat. No. 98CH36271)},
  pages        = {839--846},
  year         = {1998},
  organization = {IEEE}
}

@book{goodman2015statistical,
  title     = {Statistical optics},
  author    = {Goodman, Joseph W},
  year      = {2015},
  publisher = {John Wiley \& Sons}
}

@inproceedings{retinexnet2018,
  title   = {Deep retinex decomposition for low-light enhancement},
  author  = {Wei, Chen and Wang, Wenjing and Yang, Wenhan and Liu, Jiaying},
  journal = {arXiv preprint arXiv:1808.04560},
  year    = {2018}
}

@inproceedings{snraware2022,
  title     = {Snr-aware low-light image enhancement},
  author    = {Xu, Xiaogang and Wang, Ruixing and Fu, Chi-Wing and Jia, Jiaya},
  booktitle = {Proceedings of the IEEE/CVF conference on computer vision and pattern recognition},
  pages     = {17714--17724},
  year      = {2022}
}

@inproceedings{zeroDCE2020,
  title     = {Zero-reference deep curve estimation for low-light image enhancement},
  author    = {Guo, Chunle and Li, Chongyi and Guo, Jichang and Loy, Chen Change and Hou, Junhui and Kwong, Sam and Cong, Runmin},
  booktitle = {Proceedings of the IEEE/CVF conference on computer vision and pattern recognition},
  pages     = {1780--1789},
  year      = {2020}
}

@article{npe2013,
  title     = {Naturalness preserved enhancement algorithm for non-uniform illumination images},
  author    = {Wang, Shuhang and Zheng, Jin and Hu, Hai-Miao and Li, Bo},
  journal   = {IEEE transactions on image processing},
  volume    = {22},
  number    = {9},
  pages     = {3538--3548},
  year      = {2013},
  publisher = {IEEE}
}

@article{lime2016,
  title     = {LIME: Low-light image enhancement via illumination map estimation},
  author    = {Guo, Xiaojie and Li, Yu and Ling, Haibin},
  journal   = {IEEE Transactions on image processing},
  volume    = {26},
  number    = {2},
  pages     = {982--993},
  year      = {2016},
  publisher = {IEEE}
}

@inproceedings{mbllen2018,
  title        = {MBLLEN: Low-light image/video enhancement using cnns.},
  author       = {Lv, Feifan and Lu, Feng and Wu, Jianhua and Lim, Chongsoon},
  booktitle    = {Bmvc},
  volume       = {220},
  number       = {1},
  pages        = {4},
  year         = {2018},
  organization = {Northumbria University}
}

@inproceedings{lightennet2018,
  title     = {LightenNet: A convolutional neural network for weakly illuminated image enhancement},
  author    = {Li, Chongyi and Guo, Jichang and Porikli, Fatih and Pang, Yanwei},
  journal   = {Pattern recognition letters},
  volume    = {104},
  pages     = {15--22},
  year      = {2018},
  publisher = {Elsevier}
}

@inproceedings{retinexformer2023,
  title     = {Retinexformer: One-stage retinex-based transformer for low-light image enhancement},
  author    = {Cai, Yuanhao and Bian, Hao and Lin, Jing and Wang, Haoqian and Timofte, Radu and Zhang, Yulun},
  booktitle = {Proceedings of the IEEE/CVF international conference on computer vision},
  pages     = {12504--12513},
  year      = {2023}
}

@article{zhang2011fsim,
  title     = {FSIM: A feature similarity index for image quality assessment},
  author    = {Zhang, Lin and Zhang, Lei and Mou, Xuanqin and Zhang, David},
  journal   = {IEEE transactions on Image Processing},
  volume    = {20},
  number    = {8},
  pages     = {2378--2386},
  year      = {2011},
  publisher = {IEEE}
}

@article{yeganeh2013tmqi,
  title     = {Objective quality assessment of tone-mapped images},
  author    = {Yeganeh, Hojatollah and Wang, Zhou},
  journal   = {IEEE Transactions on Image processing},
  volume    = {22},
  number    = {2},
  pages     = {657--667},
  year      = {2012},
  publisher = {IEEE}
}

@article{cho2016peplography,
  title     = {Peplography—a passive 3D photon counting imaging through scattering media},
  author    = {Cho, Myungjin and Javidi, Bahram},
  journal   = {Optics letters},
  volume    = {41},
  number    = {22},
  pages     = {5401--5404},
  year      = {2016},
  publisher = {Optical Society of America}
}

@article{kim2023kalman,
  title     = {Three-dimensional (3d) visualization under extremely low light conditions using kalman filter},
  author    = {Kim, Hyun-Woo and Cho, Myungjin and Lee, Min-Chul},
  journal   = {Sensors},
  volume    = {23},
  number    = {17},
  pages     = {7571},
  year      = {2023},
  publisher = {MDPI}
}

@inproceedings{dong2011bcp,
  title     = {Fast efficient algorithm for enhancement of low lighting video},
  author    = {Dong, Xuan and Pang, Yi and Wen, Jiangtao},
  booktitle = {ACM SIGGRApH 2010 posters},
  pages     = {1--1},
  year      = {2010}
}

@incollection{israel1959koschmieders,
  title     = {Koschmieders theorie der horizontalen sichtweite},
  author    = {Isra{\"e}l, Hans and Kasten, Fritz},
  booktitle = {Die Sichtweite im Nebel und die M{\"o}glichkeiten ihrer k{\"u}nstlichen Beeinflussung},
  pages     = {7--10},
  year      = {1959},
  publisher = {Springer}
}

@article{he2010single,
  title     = {Single image haze removal using dark channel prior},
  author    = {He, Kaiming and Sun, Jian and Tang, Xiaoou},
  journal   = {IEEE transactions on pattern analysis and machine intelligence},
  volume    = {33},
  number    = {12},
  pages     = {2341--2353},
  year      = {2010},
  publisher = {Ieee}
}

@article{elad2002origin,
  title     = {On the origin of the bilateral filter and ways to improve it},
  author    = {Elad, Michael},
  journal   = {IEEE Transactions on image processing},
  volume    = {11},
  number    = {10},
  pages     = {1141--1151},
  year      = {2002},
  publisher = {IEEE}
}

@article{milanfar2013tour,
  title     = {A tour of modern image filtering: New insights and methods, both practical and theoretical},
  author    = {Milanfar, Peyman},
  journal   = {IEEE signal processing magazine},
  volume    = {30},
  number    = {1},
  pages     = {106--128},
  year      = {2012},
  publisher = {IEEE}
}

@book{gonzalez2009digital,
  title     = {Digital image processing},
  author    = {Gonzalez, Rafael C},
  year      = {2009},
  publisher = {Pearson education india}
}

\end{document}